\documentclass{Interspeech2024}

\interspeechcameraready 

\usepackage[table]{xcolor}
\newcommand{\cmark}{\ding{51}}%
\newcommand{\xmark}{\ding{55}}%
\usepackage{pifont}
\title{\dataset{}: A Robust Multi-accent Benchmark for Evaluating Hindi ASR Systems}

\name[affiliation={}]{Tahir}{Javed}
\name[affiliation={}]{Janki}{Nawale}
\name[affiliation={}]{Sakshi}{Joshi}
\name[affiliation={}]{Eldho}{George}
\name[affiliation={}]{Kaushal}{Bhogale}
\name[affiliation={}]{Deovrat}{Mehendale}
\name[affiliation={}]{Mitesh M.}{Khapra}

\address{
  AI4Bharat, Indian Institute of Technology Madras, India
}
\email{tahirjmakhdoomi@gmail.com, miteshk@cse.iitm.ac.in}

\keywords{non-native speech recognition, Indian accents.}

\newcommand{\dataset}[1]{\textsc{Lahaja}}

\begin{document}

\maketitle

\begin{abstract}

Hindi, one of the most spoken language of India, exhibits a diverse array of accents due to its usage among individuals from diverse linguistic origins. To enable a robust evaluation of Hindi ASR systems on multiple accents, we create a benchmark, \dataset{}, which contains read and extempore speech on a diverse set of topics and use cases, with a total of 12.5 hours of Hindi audio, sourced from 132 speakers spanning 83 districts of India. We evaluate existing open-source and commercial models on \dataset{} and find their performance to be poor. We then train models using different datasets and find that our model trained on multilingual data with good speaker diversity outperforms existing models by a significant margin. We also present a fine-grained analysis which shows that the performance declines for speakers from North-East and South India, especially with content heavy in named entities and specialized terminology. %Our code, datasets and models will be made publicly available.

   %Hindi, one of the most spoken language of India, exhibits a diverse array of accents due to its usage among individuals from diverse linguistic origins. %It is thus crucial to evaluate Hindi ASR systems on a benchmark that accurately captures this wide diversity of accents. However, none of the existing Hindi benchmarks capture this diversity. In this work, first, we address this gap by 
   %To enable a robust evaluation of Hindi ASR systems on multiple accents, we create a benchmark, \dataset{}, which contains read and extempore speech on a diverse set of topics and use cases, %from speakers from different regions across India. \dataset{} contains 
   %with a total of 12.5 hours of Hindi audio, sourced from 132 speakers spanning 83 districts of India. We evaluate existing open-source and commercially available models on \dataset{} and find their performance to be poor. We then train different models using different data sources and find that our model trained on multilingual data with good speaker diversity outperforms existing models by a significant margin. We also present a fine-grained analysis which shows that the performance notably declines for speakers from North-East and South India, especially with content heavy in named entities and specialized terminology. All the code, datasets, models and evaluation scripts developed as a part of this work will be made publicly available. \todo{Tahir: Once the abstract is done, check if it is under 1000 characters.}
\end{abstract}

\section{Introduction}

\noindent Hindi is one of the most widely spoken languages of India with 528M speakers identifying it as their first language and another 163M identifying it as their second or third language. People across the country learn and speak Hindi for personal, political and/or employment reasons, and it serves as an unofficial lingua franca for day-to-day activities in several parts of the country. %it on a day to day basis, be it courts, educational institutes, hospitals or other public service infrastructures. Globally, Hindi is spoken by 584.57 million speakers, making it the fourth most popular language in the world. India officially has recognized 48 dialectical variations for Hindi. 
As a result there is significant variation in the accents of people speaking Hindi across the country with regional influences as well as influences from the primary language. These regional influences stem from the rich linguistic diversity of India which has 22 scheduled languages, 122 major languages, and 1599 other languages, as
per the Census of 2011. Speakers of languages from the Dravidian family, like Tamil and Malayalam, showcase unique speech rhythms and ways of articulating words that stand in contrast to those from the Indo-Aryan group, including languages like Hindi, Marathi, and Gujarati. Accentual differences are also prominent within the Indo-Aryan languages, reflecting the diverse linguistic landscapes of India's northern, western, and eastern regions. Given the widespread usage and diversity, it is imperative to develop automatic speech recognition systems for Hindi which cater to multiple accents. 

\begin{figure}
    \centering
    \includegraphics[width=0.85\linewidth]{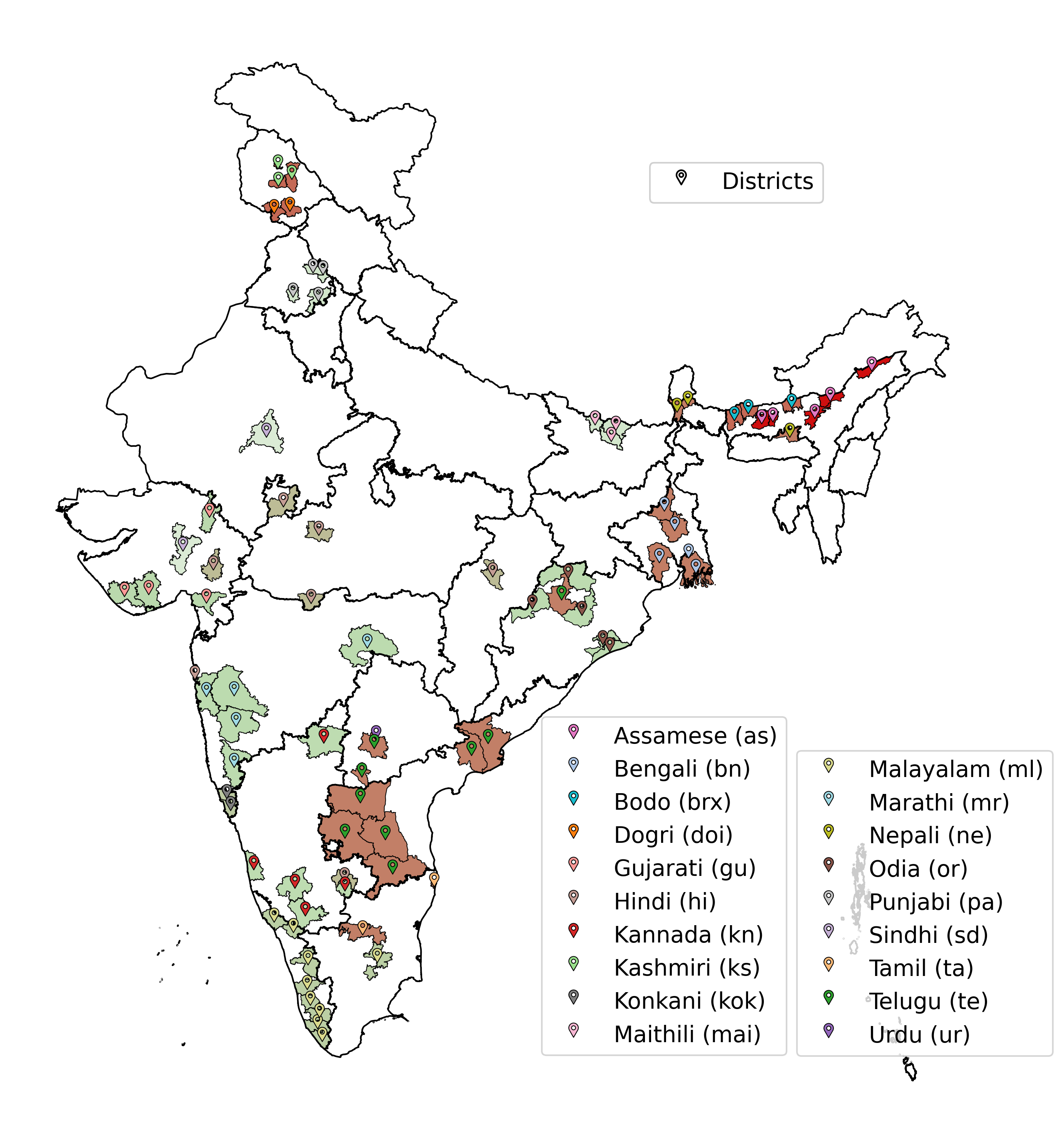}
    \caption{Different districts of India from which data was collected. The colors show how the WER of our best model varies across different regions of India (shades of green being relatively better and shades of red and brown being relatively poor).}
    \label{fig:svarah-map}
\end{figure}

While there are efforts to collect voice samples from \textit{native} speakers of Hindi \cite{indicvoices,spring,graamvaani, diwan2021multilingual,10.1609/aaai.v37i11.26521} for training ASR systems, there is no benchmark which has Hindi speakers from \textit{diverse backgrounds}, speaking with \textit{different accents}. In this work, we address this gap by releasing \dataset{}, an ASR benchmark containing multi-accent Hindi data. We follow the same collection methodology as used in \textsc{Svarah} \cite{svarah}, which is an ASR benchmark containing Indian-accent English data and \textsc{IndicVoices} \cite{indicvoices} which is an effort to collect data from native speakers only (as opposed to non-native speakers in our case). \dataset{} contains a total of 12.5 hours of Hindi data collected from 132 speakers of which 122 are non-native speakers. These speakers were spread across 82 districts spanning 18 states in India as shown in Figure \ref{fig:svarah-map}. The set of native languages spoken by these speakers encompasses 19 of the 22 constitutionally recognised languages of India, spread across 4 distinct language families. 

We evaluate existing open source and commercial models on \dataset{} to understand the current state of ASR for Hindi. In addition, we train multiple model variants based on the Conformer architecture by using different data sources. We observe that (i) our model outperforms all existing models on \dataset{} (ii) our model trained on multilingual data performs better perhaps due to better speaker diversity and (iii) in low resource monolingual settings adding synthetic code-mixed data helps. We also present a fine-grained analysis across different accents and content categories and observe that the performance is poor on speakers from North-East India and South India, with a sharp drop on content rich in named entities and terminology from specific domains. All the code, datasets, models and scripts have been made publicly available\footnote{\url{https://github.com/AI4Bharat/Lahaja}} and we hope that they will enable further research on multi-accent Hindi ASR systems.
%The current state of ASR struggle to handle accented speech due to their training data being less representative variations like accents or regional dialectical variations. The diverse phonetic and linguistic variations in accents pose significant difficulties for modern ASR systems, impacting both data collection and modelling approaches. 
%This leads to biased ASR performance across different accents, which negatively affects both ASR users and researchers. The significance of accent-robust ASR is growing in practical use cases. Substantial obstacles like sparse data availability and the absence of a standardized benchmark hinder the advancement of research. 

%In this work, we first create a robust ASR benchmark created in accent-inclusive approach from Indian subcontinent. We specifically, curate 12.5 hours of Hindi accented data across 132 speakers of which 122 speakers have Hindi as their secondary language in addition to their mother tongue. This data has been collected from 82 district of India. Second, we built a Hindi ASR system by aggregating all the publicly available data in the open-source. Third, we study the performance of our model on different accents and task categories. Fourth, we study a novel technique of mix-and-match to improve performance on English code-mixed data. All the artifacts of this work, including datasets, models and scripts will be made publicly available.

% \section{Related work}

\section{\dataset{}}
We now describe the process of creating \dataset{}. As mentioned earlier, we largely follow the same methodology as used in \cite{svarah, indicvoices} but focus on non-native speakers of Hindi. 
%We briefly describe the methodology below.

\subsection{Recruitment of Speakers}
We selected 132 participants from 18 out of the 28 states of India, of which 122 were non-native speakers and identified Hindi as their second, third or fourth language. The primary languages of these non-native speakers covered 19 of the 22 scheduled languages of India belonging to Indo-Aryan, Dravidian, and Tibeto-Burman language families. For each of the 19 languages, we recruited 3--5 participants who could speak Hindi. This included 65 males and 67 females, with 6.5 hours of male speech and 6.0 hours of female speech. We included participants from diverse age groups: 18--30, 30--45, 45--60, and 60+, with roughly equal representation in each age group. Participants came from various segments (unemployed, students, blue-collar, and white-collar) with varying education levels (upto 12th grade, undergraduates, graduates, and postgraduates). Participants were briefed about the task, and were clearly informed that their voice samples will be used to develop and evaluate speech recognition models. Their voice samples were recorded only after they willingly agreed and signed a consent form. The participants were appropriately compensated for their work according to daily wages in their region. The entire process was reviewed and approved by our Institute Ethics Committee.

\subsection{Data collection}
For recording voice samples, we used Microsoft's open-source Karya platform \cite{karya}. Once a participant is identified, we onboard them by asking them to fill a web-form which collects participant's meta-data such as age, gender, district, primary language and topics/domains of interest. Once registered, the participants are asked to download and install the Karya application. The participant then performs the following tasks on the Karya.

\noindent\textbf{Read speech:} To ensure good vocabulary coverage we use 1K sentences from Wikipedia articles covering 13 domains, as released by \cite{indicvoices}. We ask non-native speakers of Hindi to read out these sentences as it is.

\noindent\textbf{Digital interactions with voice assistants:} Following \cite{indicvoices}, we ask speakers to record utterances typically found in digital transactions with voice assistants. These digital transactions cover interactions with (i) in-home assistants for everyday tasks such as \textit{setting an alarm}, \textit{switching on the light, playing music}, etc. (ii) digital payment services covering multiple intents such as \textit{checking account balance, transferring money, paying electricity bill}, etc. (iii) online grocery shopping apps covering multiple intents such as \textit{placing an order, seeking a refund, changing delivery address, etc.} and (iv) online government services covering multiple intents  such as \textit{applying for a service}, \textit{checking the status of application}, \textit{renewing a service}, etc. The diversity in the applications covered ensures that the benchmark has a good representation of number sequences, alphanumeric codes, brand names, product names, bank names, government scheme names, application specific terminology and code mixed content (English-Hindi) typically found in such interactions. 

\begin{figure}[!t]
    \centering
    \includegraphics[width=\linewidth]{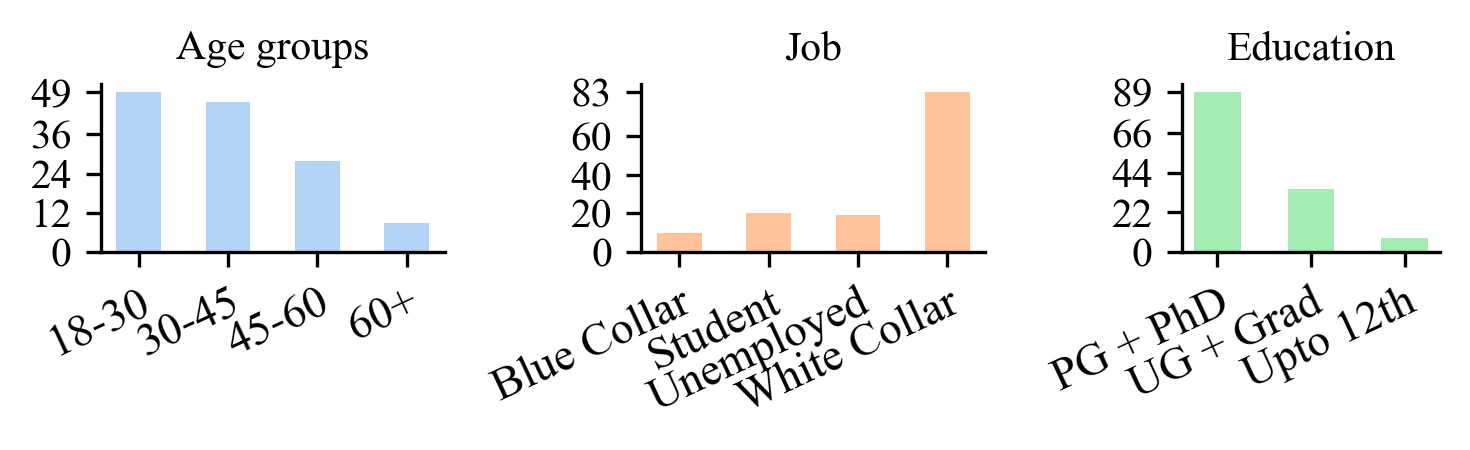}
    \caption{Demographic distribution of participants in \dataset{} across age group, job segment and educational background.}
    \label{fig:enter-label}
\end{figure}

\begin{table}[!t]
    \centering
    \footnotesize
    \caption{Statistics of \dataset{} across different native speakers. (\# Mins = \# Minutes, \# Sp = \# Speakers, \# Exclusives = \# words which are unique to that splice of data)}
    \begin{tabular}{l|cccc}
    \toprule
    \textbf{Native speakers} & \textbf{\# Mins} & \textbf{\# Sp} & \textbf{\# Words} & \textbf{\# Exclusives} \\
   % \multicolumn{1}{l|}{\textbf{\begin{tabular}[c]{@{}l@{}}Native\\Speakers\end{tabular}}} &     \multicolumn{1}{l}{\textbf{\begin{tabular}[c]{@{}c@{}}\# Min-\\utes\end{tabular}}} & \multicolumn{1}{l}{\textbf{\begin{tabular}[c]{@{}c@{}}\# Spea-\\kers\end{tabular}}} &    \textbf{\# Words} & \multicolumn{1}{l}{\textbf{\begin{tabular}[c]{@{}c@{}} \# Exclusive\\words\end{tabular}}} \\
   \midrule
    Assamese (as) & 32.6 & 6 & 1158 & 225 \\
    Bengali (bn) & 52.0 & 10 & 1786 & 471 \\
    Bodo (brx) & 43.6 & 6 & 1462 & 316 \\
    Dogri (doi) & 30.4 & 6 & 1377 & 325 \\
    Gujarati (gu) & 37.0 & 4 & 1463 & 339 \\
    Hindi (hi) & 61.1 & 10 & 2053 & 521 \\
    Kannada (kn) & 37.0 & 8 & 1367 & 294 \\
    Kashmiri (ks) & 39.5 & 7 & 1505 & 416 \\
    Konkani (kok) & 50.3 & 7 & 1432 & 280 \\
    Maithili (mai) & 34.6 & 6 & 1576 & 381 \\
    Malayalam (ml) & 38.8 & 11 & 1566 & 319 \\
    Marathi (mr) & 52.5 & 6 & 1911 & 498 \\
    Nepali (ne) & 42.3 & 5 & 1475 & 329 \\
    Odia (or) & 44.7 & 8 & 1553 & 333 \\
    Punjabi (pa) & 30.8 & 7 & 1357 & 262 \\
    Sindhi (sd) & 17.3 & 5 & 851 & 119 \\
    Tamil (ta) & 24.7 & 5 & 1030 & 201 \\
    Telugu (te) & 55.1 & 12 & 1786 & 387 \\
    Urdu (ur) & 14.8 & 3 & 727 & 114 \\
    \bottomrule
    \end{tabular}
    \label{tab:native-speakers}
    \end{table}

\begin{table}[!]
    \footnotesize
    \centering
    \caption{Statistics about different content categories in \dataset{}. (\# Mins = \# Minutes, \# Utt = \# Utterances, \# Exclusives = \# words which are unique to that splice of data)}
    \begin{tabular}{l|rcc}
    \toprule
    \textbf{Content categories} & \textbf{\# Mins} & \textbf{\# Utt} & \textbf{\# Exclusives} \\
    % \textbf{Content categories} & \multicolumn{1}{r}{\textbf{\begin{tabular}[c]{@{}c@{}}\# Min-\\utes\end{tabular}}} & \multicolumn{1}{c}{\textbf{\begin{tabular}[c]{@{}c@{}}\# Utter-\\ances\end{tabular}}} & \multicolumn{1}{c}{\textbf{\begin{tabular}[c]{@{}c@{}}\# Exclusive\\words\end{tabular}}} \\
    \midrule
    \multicolumn{4}{l}{\textbf{\textit{Read speech}}} \\
    \quad Wikipedia sentences & 41.1 & 268 & 754 \\
    \midrule
    \multicolumn{4}{l}{\textbf{\textit{Digital interactions with voice assistants}}} \\

    \quad Digital payment services & 38.2 & 263 & 188 \\
    \quad Everyday tasks & 13.5 & 257 & 144 \\
    \quad Online government services & 52.9 & 281 & 138 \\
    \quad Online grocery shopping & 32.3 & 264 & 558 \\
    \midrule
    \multicolumn{4}{l}{\textbf{\textit{Extempore conversations}}} \\
    \quad Agriculture and fisheries & 6.5 & 47 & 70 \\
    \quad Business and finance & 16.1 & 133 & 131 \\
    \quad Humanities and culture & 92.6 & 697 & 723 \\
    \quad Icebreakers & 216.6 & 1647 & 1898 \\
    \quad Leisure activities & 65.5 & 511 & 633 \\
    \quad Mass communication & 22.3 & 160 & 169 \\
    \quad Product reviews & 28.6 & 203 & 248 \\
    \quad Public resources & 35.9 & 275 & 284 \\
    \quad Science and technology & 18.9 & 153 & 181 \\
    \quad Sports and travel & 31.8 & 264 & 260 \\
    \midrule
    \multicolumn{4}{l}{\textbf{\textit{Named entities}}} \\
    \quad Task of five & 22.7 & 476 & 349 \\
    % \midrule
    % \multicolumn{4}{l}{\textbf{\textit{Keywords spotting}}} \\
    % Keywords & 3.6 & 253 & 9 \\
    \bottomrule
    \end{tabular}
    \label{tab:category-splits}
\end{table}

\noindent\textbf{Extempore conversations:} We use a carefully curated list \cite{indicvoices} of 2.5K questions from 21 domains such as tourism, government etc., and 28 topics of interest such as reading, painting etc. Next, we request each participant to select two topics they are interested in and two domains with which they are familiar and capable of answering questions about. Some sample questions include ``Technology: How have smartphones made life better?'', ``Government: Given a chance, what policies will you introduce to aid farmers in your area'' , ``Reading: Do you have a favorite book? If so, what is it and why do you like it?'' and so on. While the examples shown here are in English, the questions are translated to Hindi and shown to the participants. In addition to the above we also use some icebreaker questions to warm up the participants. These included questions about their mother tongue, their everyday life, their state/district and so on. 

\noindent\textbf{Named entities:} To get a good representation of named entities typically encountered in downstream applications, we ask users to speak any 5 numbers
%(large and small)%
, any 5 dates, any 5 person names, names of any 5 Indian cities, any 5 Indian states, any 5 Indian districts, any 5 countries, and any 5 international cities. 

Each participant thus reads 20 sentences across both read speech and digital interactions, and answers 8 questions on selected domains and topics of interest.

\subsection{Transcription}
We adhere to the guidelines as outlined in \cite{indicvoices} for transcribing the collected audio samples. We use an open-source platform, Shoonya \cite{shoonya}, for transcription which supports multiple Indian languages and a  maker-checker workflow. The workflow ensures that the initial transcript (maker) is verified by a senior transcriber (checker). All transcripts are generated in the native Devanagari script of Hindi. Our transcribers are language experts with several years of experience in transcription and translation tasks. We first split the larger audio files into segments using Silero Voice Activity Detection \cite{silero} and then provide these segments to the transcribers for transcription. Finally, we downsample the chunked audios to 16kHz, resulting in 16kHz mono 16-bit PCM wav audios.
% they needed to be split into segments first. 
% Before transcription, w
% We split larger audio files into segments using Silero Voice Activity Detection \cite{silero}. Finally, 

\subsection{Statistics}
Table \ref{tab:native-speakers} shows statistics of \dataset{} split across native speakers belonging to different languages. Table \ref{tab:category-splits} shows the statistics grouped across different content categories which allows for a fine-grained evaluation of downstream models on \dataset{}. These %content
categories were created by grouping roughly related domains and topics of interest. For example domains like education, govt., health, legal are grouped into `public resources'.
% and so on. 
% shows statistics split across different content categories in \dataset{} which would allow for a fine-grained evaluation of downstream models.
%Transcribing extempore data is a challenge because of overlapping speech, mispronounced words, dialectic differences, jargon etc. The main challenge in the transcription of accented speech data was the ambiguity caused in the meaning of the text because of the acoustic features. To transcribe the data we recruited undergraduate and postgraduate students who work on various research projects in our institute and are fluent in Hindi (having done their schooling as well a higher education in English). The transcribers were asked to verify the recorded samples to check for audio quality and to also ensure that the responses were on topic (i.e., the participants were responding correctly to the prompts). %For transcription guidelines we followed \cite{indicvoices} and created two versions of transcripts for a particular audio viz L1 and L2, L1 being faithful to the audio and L2 being a slightly normalized version of L1. 

\section{Experimental Setup}
%We now describe our experimental setup.\\
\noindent \textbf{Baselines:} To establish a baseline, we evaluate the performance of the following existing models on \dataset{}.
\begin{itemize}
    \item\textit{MMS \cite{mms}:} This is Meta's open-source 300M wav2-vec2 \cite{wav2vec2} model, supporting 1107 languages, including Hindi.
    % This is Meta's open-source model  based on the wav2-vec2 \cite{wav2vec2} architecture. It is 300M parameters multilingual model supporting 1107 languages, including Hindi.
    \item \textit{WhisperV3:} This refers to the latest open-source Whisper \cite{whisper} model, trained on 680k hours of data, having 1550M parameters and supporting 100+ languages, including Hindi
    %This is the latest version of the open-source Whisper \cite{whisper} family of models with 1550M parameters. This a multilingual model trained on 680k hours of weakly supervised data, supporting 100+ languages, including Hindi.
    \item \textit{Azure:} This refers to the Hindi speech to text systems, commercially made available by Microsoft through their SDKs. 
    \item \textit{Google Chirp:} This refers to Google USM \cite{google-usm} model which is made commercially available through Google Cloud APIs.
\end{itemize}

\noindent \textbf{IndicASR model:} We train a Conformer-L \cite{conformer} with a hybrid RNNT-CTC \cite{DBLP:journals/corr/abs-2201-05420} decoder.
%Conformer\cite{indicvoices,indicvoices} model with 120M parameters using NeMo toolkit. 
We trained 4 different variants of the model by starting with the pretrained checkpoint of an English ASR model, Nvidia-En-SSL \cite{EnSSL}, and fine-tuning it on different datasets as described below. We found that starting with english checkpoint helped in faster convergence.

\begin{itemize}
    \item \textbf{M1:} This model was trained on the Hindi subset of the \textsc{IndicVoices} dataset which contains 65 hours.
    \item \textbf{M2:} This is a multilingual model trained on the entire \textsc{IndicVoices} dataset which contains 1509 hours summed up across 22 Indian languages.
    \item \textbf{M3:} This is a monolingual model trained on 2285 hours by combining the Hindi subsets of \textsc{Vistaar} \cite{vistaar}, \textsc{Spring-Inx} \cite{spring} and \textsc{IndicVoices}.
    \item \textbf{M4:} A lot of extempore content in Indian languages is code-mixed with English, especially for non-native speakers. Hence, we do an interesting experiment where we train a model using 65 hours of Hindi data from \textsc{IndicVoices} plus an additional 65 hours of synthetic data. This synthetic data is obtained by taking 65 hours of English ULCA ASR data \cite{tulca} which contains English content spoken by Indian users. We transliterate the English transcripts to Devanagari script using an open source transliteration model, IndicXlit \cite{xlit}. Thus we created a English-Hindi \textit{mixed} dataset which contains original Hindi audios as well as English audios which are transcribed using Devanagari script (the content is in English but written in Devanagari script, as is the case in code-mixing).
\end{itemize}

\noindent We trained all the models for a maximum of 130k steps and employed early stopping with a patience of 5k steps. We set the max sequence length to 30 secs, used batch size of 16 audios per GPU on 8 GPUs with gradient accumulation of 4, resulting in an effective batch size of 512 audios. We used AdamW \cite{adamw} as the optimizer with \textit{lr} of 2.0 and Noam \cite{DBLP:journals/corr/VaswaniSPUJGKP17} as the LR scheduler. 
% \noindent We trained all the models by setting the max number of updates to 130k steps and used early stopping with a patience of 5k steps. We used a batch size of 16 audios per GPU on 8 GPUs and with a gradient accumulation of 4, resulting in an effective batch size of 512 audios. We set the the max sequence length to 30 secs. We used AdamW \cite{adamw} as the optimizer with a learning rate of 2.0 and Noam \cite{DBLP:journals/corr/VaswaniSPUJGKP17} as the learning rate scheduler. 

\noindent\textbf{Evaluation metric:} We used Word Error Rate (WER) as the metric to compare performance across models.

\section{Results and Discussion}

\noindent \textbf{Performance across models:} Referring to Table \ref{tab:model-wer}, we observe that our base model \textbf{M1} outperforms all existing models with a minimum and maximum improvement of 2.9\% WER and 13\% WER, respectively. Among the baseline models, the massively multilingual open source models perform poorly as compared to the closed source commercial models from Azure and Google. 

Next, we compare our monolingual model, \textbf{M1}, with our multilingual model, \textbf{M2}, both trained on IndicVoices. It is observed that \textbf{M2} outperforms its monolingual counterpart, by a margin of 2.7\% WER. There could be two reasons for the better performance of the multilingual model (i) on aggregate it uses much more training data than the monolingual model although the amount of Hindi data is the same (ii) it sees training data in the native language of the accents studied in this work (although from a different set of speakers). We hypothesise that the second reason is more likely as otherwise the massively multilingual MMS and Whisper V3 models which have arguably trained on much larger data would also have performed better. %performs bsThis can be attributed to cross-lingual transfer, as IndicVoices encompasses native speakers of the languages whose accents are being considered in this \dataset{}.Hence, during training in a multilingual setting, the model shares speaker characteristics across different languages, such as accent, thereby contributing to enhanced performance. 

Lastly, again referring to Table \ref{tab:model-wer}, we compare our multilingual model (\textbf{M2}) and our monolingual model (\textbf{M3}), trained using two additional sources of Hindi data: \textsc{Vistaar} \cite{vistaar}, \textsc{Spring-Inx}. Here, we clearly see the effect of adding more Hindi data and observe a further reduction of 2.4\% WER while moving from \textbf{M2} to \textbf{M3}. It would have been interesting to see the effect of training a multilingual model by combining all multilingual subsets of Vistaar, Spring and IndicVoices but due to computational constraints, we leave this as future work. %Although, this may seem counter intuitive, however the only missing piece of information here is the number of native and non-native speakers in Vistaar and Spring. We hypothesize, given the scale, diversity Vistaar and Spring, a large number of speakers should be present in the two datasets, which upon aggregation with IndicVoices actually help in further reducing the WER. 
%\todo{please write your observations here}

\begin{table}[t]
    \centering
    \caption{WERs (\%) of different models on \dataset{}, (ML = Multilingual models, \textbf{?} = Information Unavailable)}
    \footnotesize
    \begin{tabular}{@{\hspace{0.5em}}l|@{\hspace{0.3em}}c@{\hspace{0.3em}}|@{\hspace{0.5em}}c@{\hspace{0.5em}}}
    \toprule
    \textbf{Model} & \textbf{ML} & \textbf{WER\scriptsize\%} \\
    \midrule
    MMS & \cmark & 34.4 \\
    WhisperV3 & \cmark & 32.4 \\
    Azure & \textbf{?} & 28.6 \\
    Google Chirp & \textbf{?} & 22.3 \\
    \midrule
    \textbf{M1:} IndicASR {\hspace{0em}\scriptsize \textit{(Trained on IndicVoices)}} & \xmark & 19.4 \\
    % \multicolumn{1}{l|}{\hspace{2em}\scriptsize \textit{(Trained on IndicVoices)}} & & \\
    \midrule
    \textbf{M2:} IndicASR {\hspace{0em}\scriptsize \textit{(Trained on IndicVoices-Multilingual)}} & \cmark & 16.7 \\
    % \multicolumn{1}{l|}{\hspace{2em}\scriptsize \textit{(Trained on IndicVoices-Multilingual)}} & & \\
    \midrule
    \textbf{M3:} IndicASR {\hspace{0em}\scriptsize \textit{(Trained on Vistaar, Spring-Inx and}} & \xmark & \textbf{14.3} \\
    \multicolumn{1}{l|@{\hspace{0.3em}}}{\hspace{2em}\scriptsize \textit{IndicVoices)}} & & \\

    % \multicolumn{1}{l|}{\hspace{2em}\scriptsize \textit{(Trained on Vistaar, Spring-Inx and IndicVoices)}} & & \\
    \bottomrule
    \end{tabular}
    \label{tab:model-wer}
\end{table}

\begin{table}
\footnotesize
\centering
\caption{(\textbf{Left}): Effect of adding 65 hours of English-Hindi \textit{mixed} data to IndicVoices. Both models are trained in a monolingual setting. (\textbf{Right}) Comparison of performance of \textbf{M3} on native and non-native accents}
\footnotesize
\begin{tabular}{@{\hspace{0em}}l@{\hspace{0.2em}}|@{\hspace{0.2em}}c@{\hspace{0em}}}
\toprule
\textbf{Model} & \textbf{WER\scriptsize\%}  \\
\midrule
\textbf{M1:} IndicASR {\hspace{0em}\scriptsize \textit{(Trained on IndicVoices)}} & 19.4  \\
% \multicolumn{1}{l|}{\hspace{2em}\scriptsize \textit{(Trained on IndicVoices)}} & \\
\textbf{M4:} IndicASR {\hspace{0em}\scriptsize \textit{(Trained on IndicVoices,}} & 18.6 \\
% \multicolumn{1}{l|}{\hspace{2em}\scriptsize \textit{(Trained on IndicVoices,}} &  \\
\multicolumn{1}{l@{\hspace{0em}}|@{\hspace{0.2em}}}{\hspace{1.5em}\scriptsize \textit{English-Hindi mixed)}} &  \\
\bottomrule
\end{tabular}
\begin{tabular}{@{\hspace{0em}}l@{\hspace{0.2em}}|@{\hspace{0.1em}}c@{\hspace{0em}}}
\toprule
\textbf{M3} & \textbf{WER\scriptsize\%} \\
\midrule
\dataset{} & 14.3 \\
\textsc{IndicVoices} & 13.1 \\
\bottomrule
\end{tabular}

\label{tab:transliterated-ulca}
\end{table}

\noindent \textbf{Effect of adding English-Hindi \textit{mixed} data:} Referring to Table \ref{tab:transliterated-ulca}, we compare our monolingual model \textbf{M1} with \textbf{M4}, which is trained with English-Hindi mixed data. Interestingly, \textbf{M4} performs better than \textbf{M1} by $\approx$1\% WER. We hypothesize that since \dataset{} contains a significant code-mixed data, adding synthetically created code-mixed content helps when the training data is less (\textbf{M1} uses only \textsc{IndicVoices}). In a separate experiment, we found that adding synthetic code-mixed on top of resource rich settings as in \textbf{M2} and \textbf{M3} does not help.

\begin{figure}[!t]
    \centering
    \includegraphics[width=\linewidth]{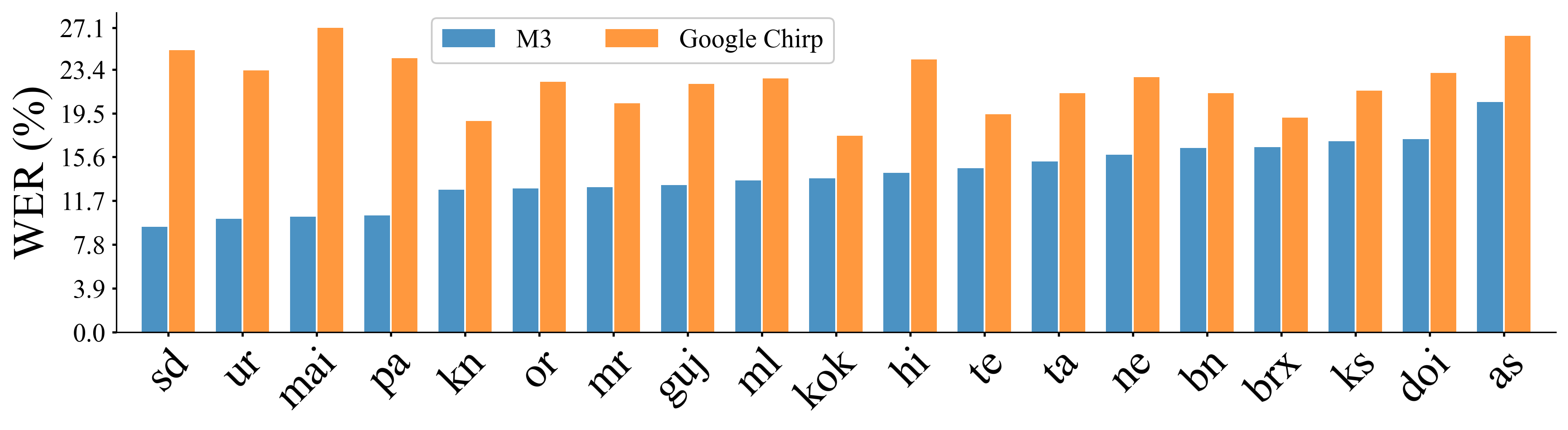}
    \caption{Performance breakdown of \textbf{M3} \& \textbf{Google Chirp} across non-native speakers.}
    \label{fig:wer-coverage}
\end{figure}

\noindent \textbf{Performance across accents:} Figure \ref{fig:wer-coverage} contains the spliced WER of our best model \textbf{M3} on different accents. It is evident that performance of \textbf{M3} decreases as we move from regions where Hindi is more popularly spoken as the $2^{nd}$ language or is closely related to the region's native language 
%being spoken in that region 
to regions where this is not the case. We observe the model performs best for languages like Urdu and Sindhi with 10.5\% WER and worst for Assamese with 20.5\% WER. More generally, from Figure \ref{fig:svarah-map} we understand that moving from Central and West India (where languages related to Hindi like Maithili, Urdu are spoken) to North East India (where languages like Assamese, Nepali are spoken) and South India (where Dravidian languages like Tamil, Telugu are spoken), we see a clear decline of performance. We hypothesise that this is due to strong influences of the primary language of the speaker which is increasingly different from Hindi. We do see surprises (e.g., we expected WER on South Indian languages, `kn' and `ml' to also be poor).

\begin{table}[!t]
    \centering
    \footnotesize
    \caption{WERs (\%) of \textbf{M3} \& \textbf{Google Chirp} across content categories of \dataset{}}
    \begin{tabular}{@{\hspace{0.5em}}lc|c|c@{\hspace{0.7em}}}
    \toprule
    \textbf{Content categories} & \textbf{Codename} &\textbf{Chirp}& \textbf{M3} \\
    \midrule
    \multicolumn{2}{@{\hspace{0.5em}}l}{\textbf{\textit{Read speech}}} \\
    \quad Wikipedia sentences & R1 & 18.6& 12.9 \\
    \midrule
    \multicolumn{2}{@{\hspace{0.5em}}l}{\textbf{\textit{Digital interactions with voice assistants}}} \\
    \quad Digital payment services &D1&40.8& 12.7 \\
    \quad Everyday tasks &D2&15.0& 13.2 \\
    \quad Online government services &D3&52.3& 13.0 \\
    \quad Online grocery shopping &D4&26.1& \cellcolor[HTML]{EEA08E}21.5 \\
    \midrule
    \multicolumn{2}{@{\hspace{0.5em}}l}{\textbf{\textit{Extempore conversations}}} \\
    \quad Business and finance &E1&17.7& 14.0 \\
    \quad Humanities and culture &E2&16.4& 13.5 \\
    \quad Icebreakers &E3&17.3& 13.3 \\
    \quad Sports and Travel &E4&19.3& 13.5 \\
    \quad Leisure activities &E5&16.9& \cellcolor[HTML]{FFF2CC}14.4 \\
    \quad Public resources &E6&18.4& \cellcolor[HTML]{FFF1CB}14.6 \\
    \quad Product Reviews &E7&19.7& \cellcolor[HTML]{FEEBC7}15.1  \\
    \quad Mass communication &E8&17.1& \cellcolor[HTML]{FBDDBD}16.2 \\
    \quad Science and Technology &E9&25.5& \cellcolor[HTML]{F6C4A9}18.4 \\
    \quad Agriculture and fisheries &E10&21.3& \cellcolor[HTML]{F0AB97}20.5 \\
    \midrule
    \multicolumn{2}{@{\hspace{0.5em}}l}{\textbf{\textit{Named entities}}} \\
    \quad Task of five &N1&63.2& \cellcolor[HTML]{E67C73}24.4 \\
    % \midrule
    % \multicolumn{2}{l}{\textbf{Keywords Spotting}} \\
    % Keywords & 7.8 \\
    \bottomrule
    \end{tabular}
    \label{tab:task-wer}
\end{table}

\noindent \textbf{Comparison with native accents:} We now compare the the performance of our model on \dataset{} and the Hindi subset of the \textsc{IndicVoices} which only contains native speakers (see the right half of Table \ref{tab:transliterated-ulca}). We observe that the performance of \textbf{M3} on IndicVoices, which consist of only native speakers of Hindi, is better than that on \dataset{}. 
%(\todo{we find a similar trend with the baseline models but due to lack of space don't report the results here - MMS and Whisper perform better on \dataset{} than IndicVoices, Azure and USM follow the same trend)}. 
This combined with the fine-grained results in Figure \ref{fig:wer-coverage} implies that \dataset{} is a good benchmark for evaluating performance across different accents. 

\begin{table}
    \footnotesize
    \centering
    \caption{Examples of errors. AC = Accent, CN = Codename}
    \begin{tabular}{@{\hspace{0.3em}}l@{\hspace{0.7em}}l@{\hspace{0.7em}}c@{\hspace{0.7em}}l@{\hspace{0.7em}}l@{\hspace{0.3em}}}
        \toprule
         \textbf{Reference} & \textbf{Predicted} & \textbf{AC} & \textbf{CN} & \textbf{Comment} \\
         \midrule
         Meghalaya & Mai kha li ya & ml & N1 & Unnecessary splitting\\
         Shonitput & Chaunitpur & ne & N1 & Confuses `sh' and `ch' \\
         Pathological & \{\} & as & E6 &  No output \\
         Glucon D & Blookandee & as & D4 & Confuses `glu' and `blu' \\
         % Aqua & \{\} & brx & E10 & No output\\
         % Species & \{\} &doi& E9 & No output\\ 
         Bade vyapaari & Body paper & or & E10 & Predicts En word instead \\
         Ansh & Aaj & ta & E9 & Predicts frequent Hi word \\
         \bottomrule
    \end{tabular}
    \label{tab:examples}
\end{table}

\noindent \textbf{Performance across different content categories:}
In Table \ref{tab:task-wer}, we present a fine-grained evaluation of the model across different content categories. The model performs well on read speech with standard vocabulary from Wikipedia, as well as everyday tasks, icebreaker questions and some domains like business and culture. The model particularly struggles in utterances which are rich in named entities (task of fives, product reviews, online grocery shopping) and in certain domains (science and technology, agriculture and fisheries) which may have very domain-specific vocabulary. We list examples of errors in Table \ref{tab:examples}.

\section{Conclusion}
We present \dataset{}, a comprehensive benchmark featuring 12.5 hours of Hindi audio from 132 speakers across 83 districts, allowing evaluation of Hindi ASR systems on multiple accents. Our evaluations reveal that existing open-source and commercial models fall short in accurately recognizing multi-accent Hindi speech, underscoring the challenge of accent diversity. However, by training models on multilingual data that encompass a broad range of speakers, we have achieved notable improvements, surpassing existing models by a significant margin. Our fine-grained analysis further emphasizes the performance gaps for speakers from North-East and South India, particularly with content laden with named entities and specialized terminology. By making our code, datasets, and models publicly available, we aim to spur further research and development of ASR systems supporting multiple accents.

\section{Acknowledgements}
We would like to thank Digital India Bhashini, the Ministry of Electronics and Information Technology (MeitY\footnote{https://www.meity.gov.in/}) of the Government of India and the Centre for Development of Advanced Computing (C-DAC\footnote{https://www.cdac.in/index.aspx?id=pune}), Pune for generously supporting this work and providing us access to multiple GPU nodes on the Param Siddhi Supercomputer. We would like to thank the EkStep Foundation and Nilekani Philanthropies for their generous grant which went into hiring human resources as well as cloud resources needed for this work. We would like to thank the team of AI4Bharat for helping us to collect data from native speakers of different languages across the country.

\bibliographystyle{IEEEtran}
\bibliography{mybib}

\end{document}

% \maketitle


\section{Supplementary material}

\begin{table}[!h]
    \centering
    \scriptsize
    \begin{tabular}{l|l}
    \toprule
    \textbf{Category} & \textbf{Task name} \\
    \midrule
    Agriculture and fisheries & DOI - Agriculture \\
     & KYP - Fishing \\
     \midrule
    Online grocery shopping & Bigbasket Commands \\
    \midrule
    Business and finance & DOI - Banking And Insurance \\
     & DOI - Business \\
     \midrule
    Digital payment services & Digital Payment Commands \\
    \midrule
    Everyday tasks & Alexa Commands \\
    \midrule
    Icebreakers & Daily Life \\
     & KYP - Basic \\
     & District Specific \\
     & GK Questions \\
     & Language Specific \\
     \midrule
    Humanities and culture & KYP - Cooking \\
     & DOI - Art \& Craft \\
     & DOI - Culture \\
     & DOI - History \\
     & DOI - Religion \\
     & KYP - Dancing \\
     & KYP - Drawing \\
     & KYP - Knitting and Stitchings \\
     & KYP - Painting \\
     & KYP - Sculpting (creating 3D art) \\
     & KYP - Singing \\
     & KYP - Writing \\
    \midrule
    Keywords & Keywords Spotting \\
    \midrule
    Leisure activities & KYP - Biking \\
     & KYP - Birdwatching \\
     & KYP - Collecting \\
     & KYP - Games \\
     & KYP - Gardening \\
     & KYP - Hiking/Trekking \\
     & KYP - Photography \\
     & KYP - Reading \\
     & KYP - Rock climbing \\
     & KYP - Running \\
     & KYP - Star gazing \\
     & KYP - Swimming \\
     \midrule
    Mass communication & DOI - Entertainment \\
     & DOI - News Media \\
     \midrule
    Product Reviews & Product Review \\
    \midrule
    Public resources & DOI - Education \\
     & DOI - Government \\
     & DOI - Health \\
     & DOI - Legal \\
     & KYP - Yoga \\
     \midrule
    Science and Technology & DOI - Geography \\
     & DOI - Stem \\
     & DOI - Technology \\
     & KYP - Astronomy \\
     & KYP - Technology \\
     \midrule
    Sports and Travel & DOI - Sports \\
     & DOI - Tourism \\
     & KYP - Traveling \\
     \midrule
    Task of Fives & Task of Fives \\
    \midrule
    Online government services & Umang Commands \\
    \midrule
    Wikipedia Sentences & Wikipedia Sentences \\
    \bottomrule
    \end{tabular}
    \caption{Caption}
    \label{tab:my_label}
\end{table}

% \bibliographystyle{IEEEtran}
% \bibliography{mybib}